# Multilingual Persuasion Detection: Video Games as an Invaluable Data Source for NLP

**Teemu Pöyhönen, Mika Hämäläinen and Khalid Alnajjar**
Faculty of Arts
University of Helsinki
teemu.poyhonen@helsinki.fi, mika.hamalainen@helsinki.fi,
khalid.alnajjar@helsinki.fi

## ABSTRACT
Role-playing games (RPGs) have a considerable amount of text in video game dialogues. Quite often this text is semi-annotated by the game developers. In this paper, we extract a multilingual dataset of persuasive dialogue from several RPGs. We show the viability of this data in building a persuasion detection system using a natural language processing (NLP) model called BERT. We believe that video games have a lot of unused potential as a datasource for a variety of NLP tasks. The code and data described in this paper are available on Zenodo.

## INTRODUCTION
We present work on constructing a multilingual persuasion annotated dataset by extracting pre-annotated dialog from video games. We use this data to build models that can detect persuasive intent automatically in dialog. Our method works on multiple languages: English, Spanish, French, Italian and German. The datasets presented in this paper have been permanently archived and released openly on Zenodo[1] to ensure replicability and reuse of our research.

The intersection between video games and natural language processing (NLP) does not seem to be fully actualized yet, as there is a lack of both game features utilizing NLP and corpora constructed from video game data. We believe there is much potential in utilizing data from video games in NLP tasks, as the data is often semi-annotated by developers through the game code or dialogue text. With this, the data is relatively accessible and eliminates the requirement for annotation work. This is crucial since the modern NLP relies heavily on supervised machine learning, which, in turn, requires substantial amounts of annotated data. In terms of video game features, creative NLP solutions could potentially, for example, improve player agency in games via players having the ability to write the dialogue for their own characters.

---

[1] https://zenodo.org/record/6341173

In the case of extracting data from video games for NLP purposes, the data is often already semi-annotated by the developers of the video games. For instance, some role-playing games (RPGs) contain programmed dialogue trees and skill or ability based dialogue options which in this sense are labeled. For example, RPGs often have a persuasion ability in dialogue which player characters can use to e.g. retrieve information or get a non-player character (NPC) to perform some action for the player character. In this sense, the dialogue option containing a persuasion attempt is either programmed into a dialogue file or can sometimes even be written into the actual dialogue text. In addition, since the dialogue sentences contain some kind of identifier, any translations and localizations that the games contain can be easily aligned for the purpose of constructing a multilingual dataset.

One can assume that all the persuasion attempts in a game generally consist of some features of persuasive language. Therefore, to show the viability of using data from video games, we have built both monolingual and multilingual classifiers to detect persuasive language. These models could either be used outside of video games to detect persuasive language, or used in a game feature in which a player is able to write their persuasion attempt instead of selecting a pre-written sentence from the dialogue. In the latter case, such a feature in a game dialogue system could potentially significantly improve both player agency as well as engagement.

Our main contributions in this work are:

1. We collect, align and clean persuasion utterances from three games and openly release the dataset.
2. We build and release monolingual and multilingual neural models for detecting persuasion automatically.
3. We evaluate the performance of the model quantitatively and qualitatively, followed by detailed analysis and discussion.

Our work fosters interdisciplinary studies between the fields of NLP and video game research as it provides mutual benefit for both of the disciplines. NLP relies heavily on the accessibility of annotated data, which is surprisingly abundant in the video game world. Video games can benefit from the latest advances in NLP in terms of technologies such as natural language interfaces to improve the overall user experience of gamers.

## RELATED WORK

There have been some takes on persuasion detection before in the field of NLP. We will describe some of the recent ones in this section, although, unlike the work presented in our paper, the previous work has been focusing mostly on persuasion detection in English. There have also been some approaches to automatic analysis of persuasion (Hidey et al., 2017; Durmus & Cardie, 2018, Shaikh et al., 2020). In particular, fake news detection has become a hot topic in the recent past (Hämäläinen et al., 2020; Oshikawa et al., 2020; Kuzim et al., 2020).

In video game research, persuasion is mainly studied from the theoretical and design point of view (Walz, 2003; Svahn, 2009; Christiansen, 2014, Ball, 2020). In the field of NLP, detection and analysis of persuasion in text has been getting a great attention lately. For instance, Dutta et al 2020 have proposed an LSTM (long short-term memory) model to identify persuasion in online discussions on the subreddit Change My View (CMV). A shared task on detection of persuasion techniques was introduced recently, SemEval-2021 Task 6 (Dimitrov, 2021). Participants in the task were requested to detect the persuasion technique, out of 22 predefined techniques, in memes (based on the image and text). The dataset was collected from Facebook groups focusing on politics, vaccines, COVID-19, and gender equality. The top two models proposed by the participants were transformer models, similar to the models we are proposing in this work. Young et al. (2011) have manually annotated a microtext corpus (600 sentences) for persuasion detection in dialog; however, it is insufficient for training neural networks and is monolingual. To the best of our knowledge, our corpus is the biggest multilingual persuasion dataset.

Video games have been slow in adapting any kind of modern NLP technology, especially Bidirectional Encoder Representations from Transformers (BERT) (Devlin, et al. 2019) models for understanding natural language. The only game (outside of educational games) utilizing NLP to our knowledge is Façade, developed by Procedural Arts (2005). In the game, the player takes an active role in a conversation by being able to input sentences and interact with two other NPCs.

Video game data has already been recently used in NLP. There has been recent work on extracting a sentiment lexicon from Skyrim (Bergsma et al., 2020). In another take, Fallout 4 dialog is extracted from the video game (Hämäläinen & Alnajjar, 2019) and then used to conduct creative dialog adaptation within the game (Alnajjar & Hämäläinen, 2019). In an example of using text classification within the context of RPGs, Kerr and Szafron (2009) utilized machine learning to detect the sophistication in dialogue lines. The system was tested using dialogue lines from Neverwinter Nights. As a disadvantage, the authors deem using film dialogue to train a classifier problematic. In this sense, when designing NLP based features in video games, it might be critical to use data from video games, even though datasets such as movie and TV-series subtitles might seem closely similar to video game texts.

## EXTRACTING THE DATA

Dialogue systems in video games are usually pre-written with players having the ability to choose different dialogue options. These options are used to then navigate dialogue trees which can vary in their complexity. Kerr and Szafron (2019) define these options with "The lowest level element of a dialogue is the dialogue line. A dialogue line may consist of any statement, question, or other similar element that the designer wishes a character to say.ue lines (p. 115) With this, player characters have the ability to act out different roles in a narrative and paths within a dialogue tree by choosing between these dialogue lines. This is a defining feature for most of the RPGs and has a significant impact on how players experience the game. In our approach, we extract in-game dialog to build a persuasion corpus.

### The video games

The data was extracted from three video games: Neverwinter Nights (Bioware, 2002), Knights of the Old Republic 1 (Bioware, 2003) and Knights of the Old Republic 2 (Bioware, 2012). Each of these games are role-playing games developed by Bioware. Neverwinter Nights is set in a fantasy world of the Forgotten Realms setting, whereas both of the Knights of the Old Republic games are set within the Star Wars universe, containing more of a science fiction setting. In these games, players can engage in dialogue with non-player characters (NPCs) in a multiple choice dialogue system and, with this, the setting is reflected in the writing of the dialogue. Additionally, a player is able to choose good and evil dialogue options in these games. In this sense, we have varied language data in terms of the settings and the choices a player can make. With this, we hope to have persuasive data represented generally, rather than dialogue from a specific setting.

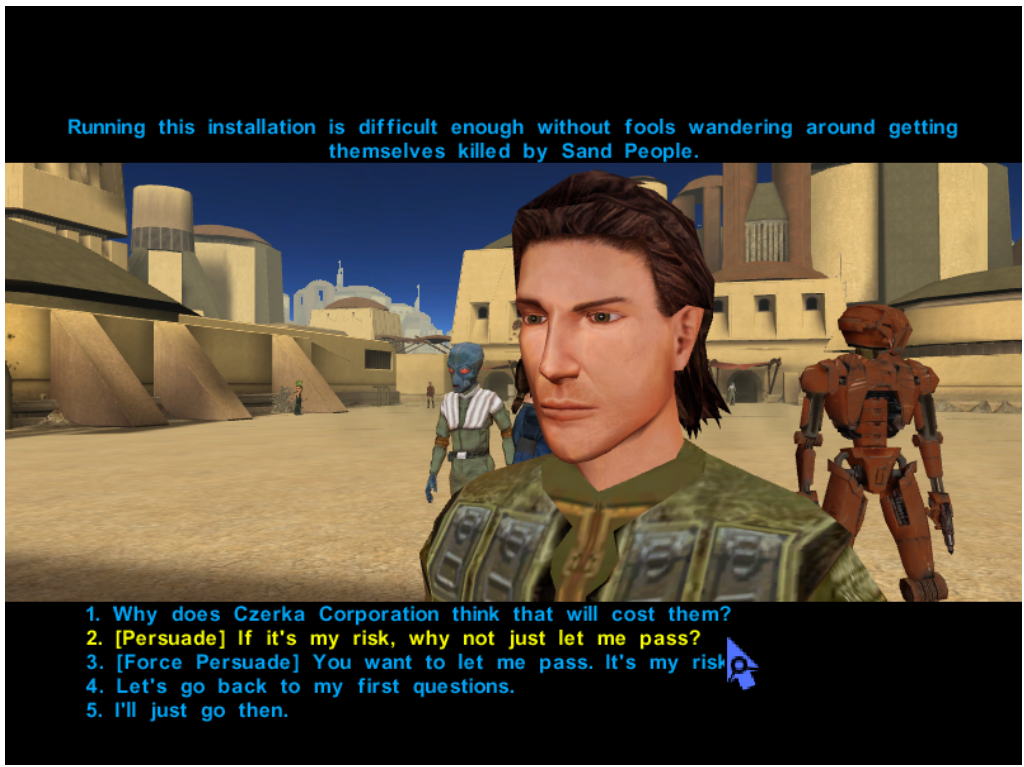

**Figure 1:** A screenshot from Star Wars: Knights of the Old Republic. In the screenshot, the *[Persuade]* option is highlighted in yellow.

The dialog in these games comes with a variety of response options that will change how the story of the game flows as seen in Figure 1. The most meaningful dialog options for us are the ones that express a persuasive intent. These options have been tagged in the game dialog files as persuasive, and we use them to gather the lines of video game dialog that express a persuasive intent. In addition, we collect other lines of dialog that are not persuasive to have samples of both persuasive lines of dialog and non-persuasive ones.

## Data extraction process

All three of these games have *dialog.tlk* files, which contain the dialogue lines of the games. These files were converted into an .xml format by using tlk2xml[2] to facilitate their processing. This also allowed for each dialogue line to be associated with a string ID. In this way, each of the English, Spanish, German, French and Italian dialogue lines were aligned by using the string ID to match the translations. This makes it possible for us to have our corpus parallelized so that we can have the same lines of dialog in the training, testing and validation for all the languages.

---

[2] https://www.mankier.com/1/tlk2xml

Each of the dialogue lines were automatically searched for any lines that contained persuasion dialogue options that the player can attempt. For this, regular expressions were utilized to find the dialogue lines which contain, for example, *[Persuasion]* within the text of the dialogue option (e.g. *[Persuade] Come on, what harm is there in telling me?*) With this, the data was labeled into *persuasion* and *non-persuasion* dialogue lines. Occasionally, the persuasion attempts can contain a hybrid tag such as *[Persuade/Lie]* in which cases the tags are simplified into *[Persuade]*. In conclusion, persuasiveness is not defined linguistically but the definition for the label emerges from the automatic extraction process. Specifically, the process marks the sentences as persuasive or not persuasive based on the method that the developers use to signal to the player that the dialogue line contains a persuasion skill challenge (including *[Persuade]* in the text of a dialogue line.)

The text was cleaned of everything except the actual dialogue text (e.g ".</string>" was removed) and made sure the UTF-8 encoding works for each of the languages. This is done to ensure consistency across all languages, and since the neural models we use expect the input to be in such a format. Additionally, some of the non-English dialogue lines contained game developer comments in English. These comments typically indicated that the dialog lines should not be translated into other languages. It is quite interesting that such lines of dialog made their way to the final published games. These were omitted from the dataset.

Sentence tokenization was applied onto each of the dialogue lines, with the respective label assigned to each of the sentences of the respective instance. We used NLTK (Bird et al., 2009) to do this tokenization. This was done to ensure that the model would not simply associate the number of sentences or the length of the input. In fact, persuasion lines tend to be only a single sentence, whereas more typical and common non-persuasive dialogue lines might even contain several sentences. We can see this when we compare a persuasive dialogue line such as *[Persuade] What other choice do you have?* with a non-persuasive dialogue line *After I overheard the deal they made with Lorso in the Czerka offices, I confronted them. When they tried to run, I chased them down and killed them.*

As a result, we have constructed a multilingual dataset where each instance is labeled as either "persuasion" or "non-persuasion," and is aligned with the translations of other languages. Before using the dataset to train our model, the instances are split into sentences (as described with sentence tokenization) with each sentence having the same label as the instance.

The resulting corpus consists of non-persuasion and persuasion instances which are each aligned with their respective translations. For each of these instances, we perform sentence tokenization. In Table 1, we present the total instances for non-persuasion sentences and persuade sentences both before sentence tokenization (as it is in the dataset) and after sentence tokenization (the training input to the BERT models.). As the number of non-persuasive sentences is greater in the corpus, we limit their number so that 80% of the dialog lines are persuasive and 20% non-persuasive. The exact numbers of sentences vary for each language after the sentence tokenization.

|  | Persuade | Non-persuade |
|---|---|---|
| English | 1572 | 6329 |
| Spanish | 1585 | 6801 |
| German | 1591 | 7011 |
| French | 1583 | 6830 |
| Italian | 1572 | 6710 |
| Multilingual (total) | 7903 | 33581 |

**Table 1:** The sizes of the constructed dataset after sentence tokenization.

We hope to demonstrate that data extracted from video games can be useful in building corpora. In this case, the data is already semi-annotated with the tag *[Persuasion]* labeling dialogue text as persuasion attempts. With this, the corpus could be useful for a persuasion detection system, or for a feature for a video game. We believe that useful data that can be extracted from video games is usually ignored and that currently there are very few published video games making use of this kind of data or NLP in general.

## PERSUASION DETECTION

In this section, we describe the method for detecting persuasive sentences of a dialogue. We base our models on BERT (Devlin et al., 2019), which is a Transformer (Vaswani et al., 2017) based multi-purpose neural language model. We compare the performances of different models, especially those based on multilingual BERT and separate BERT models for each individual language.

### The models used

Our approach for classifying persuasive sentences in this paper is to fine-tune pre-trained BERT models using transfer learning. BERT models have recently provided state-of-the-art results for natural language understanding and classification. We present 11 models in total. Specifically, five models are multilingual BERT models (Devlin et al., 2019) which have been finetuned with sentences from one of the languages (English, Spanish, German, French and Italian.) We have also trained the multilingual BERT model using the entire dataset with all of the languages. Additionally, we have fine-tuned five BERT models for each language. The models are by Schweter (2020) for Italian and German, CamemBERT by Martin et al (2020) for French and BETO: A Spanish BERT model by Cañete, et al. (2020) for Spanish. The English BERT model used for fine-tuning is the English only model presented by Devlin et al. (2019).

The models are fine-tuned to take a persuasive sentence as input and to predict its label (persuasive/non-persuasive) as output. The models are trained for 5 epochs with 500 warm-up steps and 0.01 weight decay. We set a fixed random seed of 42 to ensure replicability of the results. We use the Transformers Python library (Wolf et al., 2020) in our experiments. In particular, we use the sequence classification model. This is a Bert transformer with a sequence classification head on top of it, i.e. it has a linear layer on top of the pooled output. Dropout (Srivastava et al, 2014) is applied on the pooled output with a probability of 10% to reduce overfitting. This is important since the majority of samples in our corpus represent non-persuasive text, which could easily lead into the model only predicting non-persuasive labels without dropout. The linear layer has two outputs, and the final prediction is achieved by applying softmax and returning the highest score as the prediction for persuasion.

The corpus is split evenly on 70% of the dialog lines for testing, 15% for training and 15% for validation. The splits are shared across the languages and the splits are the same in all of our experiments. The actual number of sentences in each split for each language varies slightly because the translators have adapted different translation strategies. For example, the English *You shall find I am always very committed - to gold.* and its German translation *Ihr werdet feststellen, dass ich jeder Sache sehr ergeben bin... wenn die Bezahlung stimmt.* would result in different sentence tokenization as the German translation uses three dots instead of a hyphen. It is also to be noted that the translations are not always fully equal to the source language.

## RESULTS AND QUANTITATIVE EVALUATION

In this section, we present the quantitative results for all of the models and evaluate their performance metrics. For each model, we report the overall accuracy. We also report the F1, recall and precision scores for both of the labels of all the labels, as well as the macro average F1 and weighted average F1 scores for all the models.

In Table 2, we can see the results for the multilingual BERT model. The language specific columns show how well the model performed in the test set of a given language when the model was fine-tuned with the training data of only that language. The final column indicates how well the model that was trained on all the languages performed in the test set consisting of all the languages. All of the models performed rather poorly in correctly identifying persuasive sentences, although the model trained on all the languages achieved the highest overall results. The poor performance can be explained by the fact that the multilingual models are trained on Wikipedia, which is, as a domain, quite far away from the type of text one might encounter in video game dialog.

| Multilingual BERT | English | Italian | German | French | Spanish | All languages |
|---|---|---|---|---|---|---|
| Accuracy | 0.76 | 0.61 | 0.61 | 0.81 | 0.79 | 0.52 |
| Macro avg.(F1) | 0.51 | 0.50 | 0.5 | 0.45 | 0.51 | 0.51 |
| Weighted avg. (F1) | 0.72 | 0.65 | 0.65 | 0.73 | 0.74 | 0.56 |
| **persuade** | | | | | | |
| Precision | 0.25 | 0.20 | 0.20 | 0.20 | 0.32 | 0.28 |
| Recall | 0.11 | 0.34 | 0.38 | 0.00 | 0.09 | 0.96 |
| F1- score | 0.15 | 0.26 | 0.26 | 0.01 | 0.13 | 0.43 |
| **no_persuade** | | | | | | |
| Precision | 0.81 | 0.81 | 0.83 | 0.82 | 0.81 | 0.98 |
| Recall | 0.92 | 0.68 | 0.66 | 1.00 | 0.96 | 0.42 |
| F1- score | 0.86 | 0.74 | 0.74 | 0.90 | 0.88 | 0.59 |

**Table 2.** Performance metrics for the multilingual BERT models.

| Monolingual BERT | English | Italian | German | French | Spanish |
|---|---|---|---|---|---|
| Accuracy | 0.87 | 0.89 | 0.88 | 0.90 | 0.90 |
| Macro avg.(F1) | 0.79 | 0.82 | 0.78 | 0.82 | 0.83 |
| Weighted avg. (F1) | 0.87 | 0.89 | 0.87 | 0.89 | 0.90 |
| **persuade** | | | | | |
| Precision | 0.68 | 0.73 | 0.68 | 0.78 | 0.77 |
| Recall | 0.66 | 0.70 | 0.59 | 0.63 | 0.68 |
| F1- score | 0.67 | 0.71 | 0.63 | 0.70 | 0.73 |
| **no_persuade** | | | | | |
| Precision | 0.92 | 0.93 | 0.91 | 0.92 | 0.93 |
| Recall | 0.92 | 0.94 | 0.94 | 0.96 | 0.95 |
| F1- score | 0.92 | 0.93 | 0.93 | 0.94 | 0.94 |

**Table 3.** Performance metrics for the monolingual BERT models.

Table 3 presents the results for the several different monolingual BERT models that were fine-tuned with the training data of the same language as the model. The results are rather good for both labels and all of the languages. This can be explained by the fact that monolingual BERT models were trained on a variety of different domains of data, which facilitates the fine-tuning as the number of out-of-vocabulary words decreases.

The main difference in the multilingual and monolingual BERT models is that the multilingual models seem to quantitatively perform rather poorly on predicting the positive class (persuade.) In this way, this decreases the performance of the multilingual BERT models into rather low levels, with the macro average F1 score being close to 0.5 for all the models. In most of the models, the performance on the negative class (no_persuade) is high, yet this is to be somewhat expected as it is the majority class.

The performance metrics for the monolingual BERT models are much higher than their multilingual counterparts. The macro average F1 score averages to 0.8 and the weighted average F1 score to 0.9 across all the models. This can be attributed to the fact that the accuracy of the positive class is much higher. In fact, the accuracy is at a level that indicates the fact that the models have been successfully trained to recognize patterns from the set of data.

**Error Analysis**

By looking at the predictions made by the models, we can see that while the models can perform relatively well at classifying persuasive language, they often fail to detect indirect meanings beneath the surface level meaning. In this section, we will look at the results from the English, Spanish and French models leaving out the German and Italian one due to our own limited knowledge in these languages.

There are some instances in which the model has predicted a non-persuasion dialogue option as persuasive. Nevertheless, the model might not be entirely incorrect, as even though these parts of dialogue were not written to use the persuasion ability, the sentences could in fact be argued to be persuasive. For instance, in a sentence such as *Peragus was nothing compared to what I'll do if you don't give me back my ship.,* we can see that the speaker is trying to convince the listener to do something (give their ship back) and the instance could be considered as a persuasive dialogue line. Even though the aforementioned sentence is persuading through the means of intimidation, there are many examples of sentences that use intimidation and are also tagged to be persuasive language.

We can see the above types of false positives as evidence that the model has learned to detect persuasive language to some extent. In this way, while there exist some sentences in the game that do not contain a *[Persuade]* tag and which human annotators would likely label as persuasive, the models seem to also agree with us that the sentence is persuasive. In fact, we can see this in the sentence *Work for Czerka, and be handsomely rewarded.,* which we could argue to be persuasive language and which the model has predicted as persuasive.

The model is able to detect quite well most non-persuasive sentences, as these sentences are commonly more descriptive rather than the player character or an NPC requesting something. For instance, *A technician named Chano, located in unit 2B in Residential Module 082, handles Czerka's droid maintenance.* and *Are you aware that* *the white dragon is the weakest of the many draconic species?* could be considered as sentences that are not too challenging for the model to detect as non-persuasive sentences.

Sometimes, persuasion attempts are not direct, but the player might be given a dialogue option in which they indirectly hint towards something. On the surface, the direct meaning might seem like a regular statement, but considering the situation and context we understand the sentence to mean something more indirectly. We can see examples of this in a sentence such as *Don't worry, I won't tell anybody your secret.* and also in the sentence *You didn't see anything.*

The model can predict a persuasion sentence reasonably well, especially when there is a mention of some type of currency, e.g. gold or credits (*I'll give you 500 gold.*) Similarly, if there is a direct command (*Give me the baby or die, now!*) or some mention of e.g. trust or a promise, (*You can trust me.*) the models usually can predict the sentence to be a persuasive one.

We can observe similar tendencies in the Spanish and French test data as well. The Spanish model has correctly identified many persuasive questions such as *Confía en mí, ¿vale?* (Trust me, ok?) and *¿Tengo aspecto de jedi?* (Do I look like a jedi?). This has also led to many false positives where a non-persuasive interrogative sentence looks similar to a persuasive one, for example, *¿Seguro que era para hoy?* (Are you sure it was for today?) and *¿Significará algo nuestro sacrificio?* (Will our sacrifice mean anything?). The cases where the model produced false negatives, i.e. predicted persuasive sentences as non-persuasive, are mostly cases where the sentences are ambiguous and their persuasiveness could only be resolved in a wider context, such as *Te quiero, Valen* (I love you, Valen) and *¿Qué ocurrió?* (What happened?).

There is a significant overlap on what the French and Spanish models produced. Several questions are identified correctly as persuasive such as *Cela mérite bien un peu plus, non ?* (It is well worth a bit more, right?) and *On est copains, non ?* (We are friends, right?). As observed before, the models tend to classify sentences containing numbers as persuasive, this can also be seen in the French results as a source of false positives such as *Pour 1000 crédits, je vous trouve un moyen d'accès* (For 1000 credits, I will find you a way to get in). We can also see questions in the list of false positives such as *N'est-il pas possible de contacter l'infirmerie ?* (Is it not possible to contact the hospital?) and *Pourquoi en avez-vous besoin ?* (Why do you need that?). In the false negatives, we can see plenty of ambiguous sentences as well such as *Bien sûr que vous le pouvez* (Of course you can) and *Je vous attends ici* (I wait for you here).

## DISCUSSION AND CONCLUSION
For detecting persuasive sentences, our models performed reasonably well, especially the monolingual BERT models. We believe simply adding data from more games and instances of persuasive dialogue can improve the performance of these models significantly. This would require some normalization, since in different games the persuasion skill or ability might exist under some different label, e.g. *charisma*. In this sense, it is sometimes not entirely clear which dialogue options count as persuasive language in the same way between two games. For instance, while the games discussed in this paper contain *[Persuade]* and *[Persuade/Lie]* as tags, some other games might have narrower abilities related to persuasion, such as *intimidation* or *seduction*. In this sense, while some of the persuasive sentences in our data set could be considered as either intimidation or seduction, it depends on the set of video games how the sentences should be marked.

The described phenomenon where a persuasive sentence which actually is not tagged with *[Persuasion]*, yet the models detect the persuasiveness of the sentence (hence giving a false positive prediction) is something that might be addressed in future work. For instance, one could utilize the predictions by examining the false positives and relabeling those instances in the dataset. This would most likely improve the performance of the models.

It is interesting to note that the three languages we analyzed further in the error analysis section showcase very similar results even though the original training data for the BERT models had been different for each language and the sentences in our corpus are not exact word to word translations. This might also highlight the limitations of how good of results we can expect from a BERT based model given that similar phenomena caused similar issues across languages.

The current work we have presented in this paper has been focusing on detecting persuasive intent on a sentence level. As we saw in our error analysis, some of the sentences are too ambiguous and might very well be sincere rather than persuasive. The key to the persuasive interpretation of these particular sentences lies in the context. Contextual factors are not always easy to be included into a BERT model or any other modern NLP model, because context is such a wide concept. Too much context will make the training data too noisy and the model cannot learn anything, and too little context will not provide the model with enough information to improve on the prediction of the sentences that rely on context.

There are types of metadata that could potentially improve these models. For example, if it is possible to extract the previous sentences and the sentences after the dialogue option, it would be possible to input them as context into an NLP model. Video games are also multimodal in the sense that they contain visual information and sometimes even spoken audio. Multimodal NLP research has only recently started to gain momentum and, in the future, it might be useful to apply multimodal models into this task as well. This is especially true for the most modern AAA games where facial gestures are more lifelike and nuanced. These gestures could provide information about persuasive intent.

In this paper, we have built a novel dataset for the task of persuasion detection. To the best of our knowledge no such data exists for languages other than English. The fact that our data is parallel makes it possible to conduct more research in the future on persuasive intent and its realization across different languages. In order to make further research possible, we have released the data openly on Zenodo[3] for other researchers to use.

The multilingual BERT model did not work too well, even when it was trained with the training data of all of the languages. In theory, more data that is essentially paraphrased from the English data should lead to a performance gain due to data augmentation. However, the gain was very modest in comparison with the monolingual BERT models. We believe that this is not a question of multilingualism and monolingualism, but rather a question of the textual domain the training data represented. Wikipedia, which is the basis of the multilingual BERT model, can only provide the model with encyclopedic knowledge which might be useful for tasks that rely only on semantics. Persuasion, however, is a purely pragmatic phenomenon and it requires a richer set of textual domains to be captured correctly.

We hope to have demonstrated the viability of using data from video games in an NLP system. In this paper, we have extracted semi-annotated data from three RPGs, and besides some data processing have not done any annotation ourselves. Nevertheless, we have fine-tuned BERT models to classify persuasiveness with reasonable results. Lastly, we hope to witness the intersection between NLP and video games grow in the future.

## ACKNOWLEDGEMENTS

This work was financed by the Society of Swedish Literature in Finland with funding from Enhancing Conversational AI with Computational Creativity.

---

[3] https://zenodo.org/record/6341173